\documentclass[letterpaper, 10 pt, conference]{ieeeconf}  

\IEEEoverridecommandlockouts                              

\usepackage{stfloats} 

\usepackage{graphics} 
\usepackage{epsfig} 
\usepackage{mathptmx} 
\usepackage{times} 
\usepackage{amsmath} 
\usepackage{amssymb}  

\usepackage{hyperref}
\usepackage{algorithm}
\usepackage{algpseudocode}
\usepackage{multirow}
\usepackage[table]{xcolor}
\usepackage{lipsum}
\usepackage{comment} 
\usepackage{pifont}
\usepackage{listings}
\usepackage{bm}
\usepackage{footmisc}
\usepackage{balance}
\usepackage{cite}

\title{\LARGE \bf
Real-Time Operator Takeover for Visuomotor Diffusion Policy Training
}
\author{Marco Moletta$^1$, Michael C. Welle$^{1,2}$, Nils Ingelhag$^1$, Jesper Munkeby$^1$, Danica Kragic$^1$
\thanks{1-KTH Royal Institute of Technology, Sweden, {\it\small \{moletta, mwelle, ingelhag, munkeby, dani\}@kth.se}}
\thanks{2-INCAR Robotics AB, Sweden, {\it\small michael.welle@incar-robotics.se}  }
}

\begin{document}

\maketitle

\thispagestyle{empty}
\pagestyle{empty}

\begin{abstract}
We present a Real-Time Operator Takeover (RTOT) paradigm that enables operators to seamlessly take control of a live visuomotor diffusion policy, guiding the system back to desirable states or providing targeted corrective demonstrations. Within this framework, the operator can intervene to correct the robot’s motion, after which control is smoothly returned to the policy until further intervention is needed. We evaluate the takeover framework on three tasks spanning rigid, deformable, and granular objects, and show that incorporating targeted takeover demonstrations significantly improves policy performance compared with training on an equivalent number of initial demonstrations alone. Additionally, we provide an in-depth analysis of the Mahalanobis distance as a signal for automatically identifying undesirable or out-of-distribution states during execution. Supporting materials, including videos of the initial and takeover demonstrations and all experiments, are available on the project website\footnote{\label{fn:website} \url{https://operator-takeover.github.io/}}.

\end{abstract}

\section{Introduction}
Imitation learning (IL) has shown strong potential for automating complex robotic manipulation tasks in both domestic and industrial settings. Recent successes include cooking and wiping~\cite{fu2024mobile}, serving food and opening bottles~\cite{ingelhag2024robotic}, as well as industrial tasks such as packaging and object assembly~\cite{qi2026llm}, and deformable object manipulation in surgical automation~\cite{scheikl2024movement}. Beyond these demonstrated applications, IL has also been highlighted as a promising direction for increasing automation in the textile industry, where deformable object handling remains difficult to engineer manually~\cite{longhini2025unfolding}. These results suggest that IL offers a practical route to higher levels of robotic automation without explicit task modeling or reward engineering. 

A central ingredient of modern IL methods is the availability of expert demonstrations. Performance therefore depends strongly on the quality and coverage of the collected data. Existing efforts have focused either on improving demonstration quality after collection, for example through noise reduction~\cite{wang2023imitation}, or on enabling more natural and precise teleoperation through interfaces such as virtual reality controllers or hand tracking~\cite{moletta2023virtual, qin2023anyteleop}, augmented reality~\cite{van2024puppeteer, chen2024arcap}, or leader-follower puppeteering approaches~\cite{shaw2024bimanual, yang2024ace}. However, better teleoperation alone does not address a more fundamental limitation of imitation learning: collecting demonstrations that cover the failure modes encountered at deployment time. Policies trained only on successful expert rollouts often perform well in distribution, yet degrade when they drift into unfamiliar states for which no recovery behavior was demonstrated. In practice, such failures are difficult to anticipate during initial data collection, even for experienced operators.

\begin{figure} \centering \includegraphics[width=0.99\linewidth]{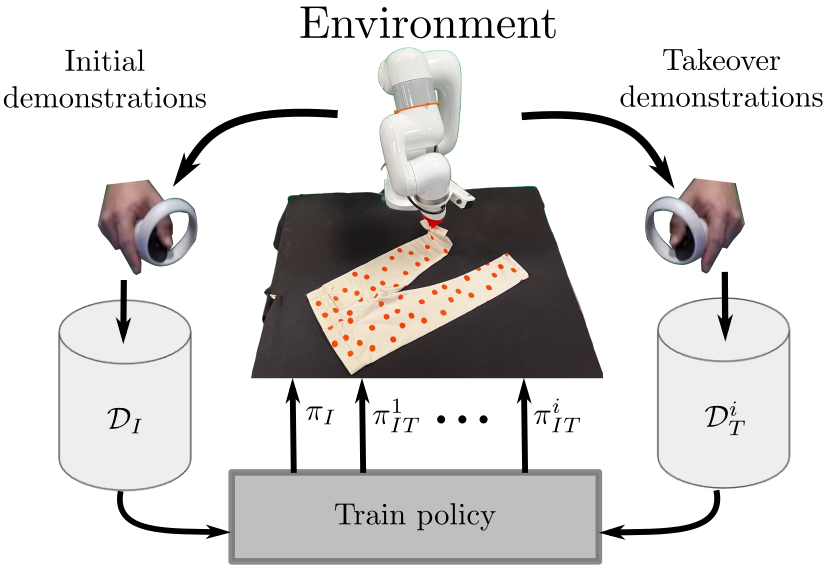} \caption{Real-Time Operator Takeover paradigm: after training a policy on a small set of initial demonstrations, the policy is deployed in the environment while the operator remains on standby. When the policy enters an undesirable state, the operator seamlessly takes over control, and only the takeover segment is recorded as an additional demonstration. The policy is then retrained with this targeted data, and the process can be repeated until the desired level of performance is reached.} \label{fig:fig1} \end{figure}

In this work, we address this limitation with a \emph{Real-Time Operator Takeover} (RTOT) paradigm. Rather than attempting to anticipate failure cases in advance, RTOT collects corrective demonstrations precisely when they occur. We first train an initial policy from a relatively small set of demonstrations and deploy it in the real environment while the operator remains on standby. When the robot enters, or is about to enter, an undesirable state, the operator seamlessly takes control, guides it back to a desirable state, and then returns control to the policy. Only the takeover segment, together with the immediately preceding context, is recorded as new training data. The policy is then retrained on the union of the initial and takeover demonstrations, and the process can be repeated iteratively.
The overall framework is illustrated in Fig.~\ref{fig:fig1}. This design has three main advantages. First, it targets precisely the states that the current policy cannot handle, instead of spending additional effort on behaviors the policy already performs reliably. Second, it turns deployment failures into supervision, making data collection adaptive to the policy’s weaknesses. Third, it reduces operator burden, since the human intervenes only when needed and only for a short corrective segment rather than re-demonstrating full trajectories. RTOT therefore provides a practical way to improve robustness while maintaining demonstration efficiency.
We evaluate RTOT on two different robot platforms and three real-world manipulation tasks spanning rigid, deformable, and granular objects. Our results show that takeover demonstrations substantially improve policy performance compared with training on an equivalent number of initial demonstrations alone, despite being shorter and more targeted. We further analyze the Mahalanobis distance as a signal for identifying undesirable or out-of-distribution states during execution, and show that it provides useful insight into critical failure points encountered by the policy.

Our contributions are as follows:
\begin{enumerate}
    \item We introduce the Real-Time Operator Takeover (RTOT) paradigm for training visuomotor diffusion policies through targeted human intervention during deployment, converting failure cases into corrective demonstrations for subsequent training.
    \item We demonstrate across three tasks involving rigid, deformable, and granular objects, using two robot setups and two camera placements, that using RTOT improves policy robustness and performance while relying on shorter, more targeted demonstrations.
    \item We analyze the Mahalanobis distance as a signal for identifying undesirable or out-of-distribution states during execution, providing novel insights on its applicability for OOD detection.
\end{enumerate}
\section{Background \& Related Work}
We organize the related work around visuomotor diffusion policies, imitation learning with a focus on out-of-distribution detection, and human-in-the-loop learning systems.

\subsection{Visuomotor Diffusion Policies}

Diffusion models~\cite{ho2020denoising}, originally introduced in generative image modeling, have recently become popular in robotics and have shown strong generalization compared with traditional discriminative models~\cite{li2023onthe}. Fundamentally, visuomotor diffusion policies iteratively transform a known prior, such as Gaussian noise, into task-relevant action sequences by learning a stochastic map to the target action distribution.

A key advantage of diffusion policies is their ability to learn effective behaviors from relatively few demonstrations~\cite{chi2023diffusion}, while requiring little explicit task or environment modeling. They have been applied to planning~\cite{kapelyukh2023dallebot, sridhar2024nomad}, robotic manipulation~\cite{reuss2023goal}, mobile bi-manual manipulation in household settings~\cite{fu2024mobile}, and deformable object manipulation in robot-assisted surgery~\cite{scheikl2024movement}. Given these properties, the quality of demonstrations becomes especially important for training effective diffusion policies.

\subsection{Imitation Learning and Out-of-Distribution Detection}

Imitation learning enables robots to acquire skills from expert demonstrations~\cite{pomerleau1998alvinn}, and recent work has improved its generalization through historical context~\cite{shafiullah2022behaviour}, alternative objectives~\cite{florence2022implicit}, multi-task and few-shot learning~\cite{dasari2021transformers}, and language conditioning~\cite{shridhar2023perceiver}.

A persistent challenge, however, is compounding error~\cite{ross2010efficient}, which can drive policies into difficult-to-recover out-of-distribution (OOD) states~\cite{ross2011reduction, tu2022sample}. Prior work has addressed this through on-policy expert corrections such as DAgger and its variants~\cite{ross2011reduction, kelly2019hgdagger, menda2019ensembledagger}, reward shaping or conservative objectives~\cite{kumar2020conservative}, synthetic offline data generation~\cite{zhou2023nerf}, and recovery policies~\cite{reichlin2022manifold}. Despite progress in OOD detection~\cite{yang2024generalized, sinha2022system}, diffusion policies generally rely on demonstration coverage rather than explicit mechanisms for detecting and/or correcting distribution shifts during deployment. Our work instead focuses on providing a framework for seamless human intervention at such failure points and enable targeted data collection.

\subsection{Human-in-the-Loop and Real-Time Takeover}

Human-robot interaction (HRI) has advanced rapidly in recent years, reshaping how humans and robots collaborate~\cite{obaigbena2024ai}. Much of this progress has been driven by machine learning and the integration of multimodal data~\cite{wang2024multimodal}. In robotic manipulation, teleoperation plays a central role by combining human cognitive skills and task expertise with the physical capabilities of robotic systems~\cite{darvish2023teleoperation}.

A prominent class of teleoperation methods relies on virtual, augmented, and mixed reality (VAM) interfaces, which provide a shared interaction space between human and robot. These frameworks have been used in areas such as motion planning and HRI~\cite{makhataeva2020augmented}, as well as for controlling robot manipulators through position or velocity commands~\cite{xu2022shared}. Applications include object handover~\cite{ortenzi2022robot}, visualization of robot intentions in delivery tasks~\cite{chandan2021arroch}, and cloth manipulation~\cite{moletta2023virtual}. However, to the best of our knowledge, existing HRI frameworks for robotic manipulation do not explicitly support seamless human takeover for corrective intervention while simultaneously recording the corrective segment for retraining. Related ideas have appeared in takeover request (TOR) frameworks for autonomous driving~\cite{lindemann2019exploring} and in manipulation systems with action correction~\cite{moletta2023virtual, ma2025human, spencer2020learning}, but they do not combine seamless takeover with the collection of new corrective demonstrations. This is precisely the setting addressed by our RTOT paradigm.
\begin{figure*}[ht]
    \centering
    \includegraphics[width=0.99\textwidth]{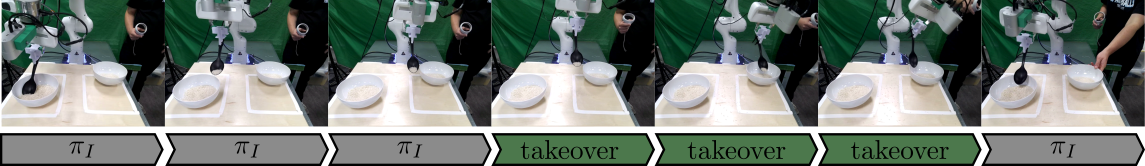}
    \caption{Real-Time Operator Takeover during policy execution (\textit{Rice scooping} task): the initial policy $\pi_I$ controls the robot until it reaches an undesirable state or a state from which failure is likely. The operator then seamlessly takes over (fourth frame), executes a corrective segment, and returns control to $\pi_I$.}
    \label{fig:takeover_frames}
\end{figure*}

\section{Methodology}
\label{sec:methodology}

We now describe our framework for real-time operator takeover for visuomotor diffusion policy training. The core idea, illustrated in Fig.~\ref{fig:takeover_frames}, is to first collect an initial dataset of demonstrations, $\mathcal{D}_I$, and use it to train an initial policy, $\pi_I$. The policy is then deployed while the operator remains on standby. When the system approaches an undesirable state, the operator can seamlessly take control, provide a short corrective intervention, and return control to the policy. These takeover segments are recorded as new demonstrations, and a new policy $\pi_{IT}^1$ is trained on the union of the initial and takeover data. This process can be repeated iteratively, allowing the dataset to expand around the failure cases actually encountered during execution rather than relying solely on the initial demonstrations.

\subsection{Real-Time Takeover}

A visuomotor diffusion policy~\cite{chi2023diffusion} is initially trained on a demonstration dataset $\mathcal{D}_I$. Demonstrations are collected by teleoperating the robot with Quest2ROS~\cite{welle2024quest2ros}, which relays VR controller velocities to a Cartesian velocity controller. The operator controls the robot by pressing the side trigger button, and only the actions executed while the trigger is pressed are recorded as $6$D end-effector velocities. Idle periods, such as pauses for hand repositioning or stopping the robot, are omitted from the recorded data.
After training, the initial policy $\pi_I$ is deployed on a robot, generating $6$D velocity commands. During execution, a ring buffer continuously stores recent observations, including the RGB camera images and proprioceptive information (the robot's pose). If the operator observes that the policy is moving toward an undesirable state, they press the trigger button to intervene, immediately overriding the policy with teleoperated commands.

\begin{figure}[t]
    \centering
    \includegraphics[width=0.99\linewidth]{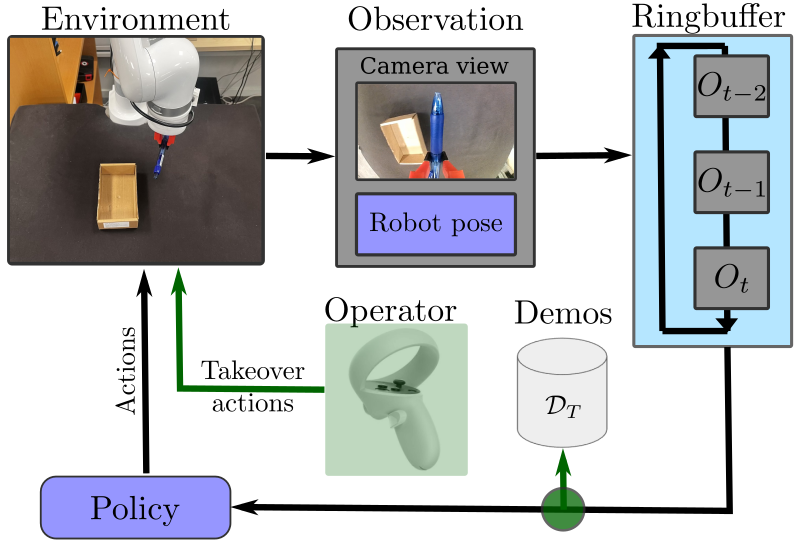}
    \caption{Illustration of the Real-Time Takeover process (\textit{Pen-in-box} task). During policy execution, observations are continuously stored in a ring buffer. When the operator takes control using the VR controller, the buffered observations together with the subsequent corrective segment are recorded as a new demonstration in $\mathcal{D}_T$.}
    \label{fig:takeover}
\end{figure}

To preserve the context leading to the intervention, the contents of the ring buffer, which capture the observations immediately preceding the intervention, are also saved as part of the new takeover demonstration. Subsequent observations and actions are then recorded while the operator guides the system back to a desirable state. Once the correction is complete, releasing the trigger seamlessly returns control to the policy. This procedure produces compact and targeted data focused on the states the current policy cannot handle reliably. Rather than collecting additional full demonstrations, the operator only contributes short corrective segments, which are then used to retrain the policy and improve robustness to failure cases encountered during deployment. A schematic of this process is shown in Fig.~\ref{fig:takeover}.
By iteratively adding takeover demonstrations, our RTOT paradigm ensures that the training set expands to include recovery behaviors and other difficult states missing from the initial dataset, resulting in a more robust policy.

\section{Experiments}
The primary goal of our experiments is to evaluate whether real-time operator takeover improves visuomotor policy performance across three tasks involving granular, rigid, and deformable objects using two robot setups.

\subsection{Experimental Setup} 
The first setup involves a Franka Emika Panda robot with an Orbbec Femto Bolt RGB camera mounted on the end effector. The second involves a Ufactory Lite6 with an OAK-D Pro W RGB end-effector camera and an Orbbec Femto Bolt RGB-D scene camera. We evaluate \textit{Rice Scooping} on the Franka using only end-effector observations, and \textit{Pen-in-box} and \textit{Trousers Folding} on the Lite6 using both end-effector and scene observations (separately), allowing us to assess the effect of different camera views on task performance and on the takeover paradigm. The tasks and success criteria are described below.

\textbf{Rice Scooping}: The goal of the task is to transfer as much rice (measured in grams) as possible from a full bowl to an empty bowl within $45$ seconds. The receiving bowl can be placed anywhere within a predefined rectangular region. Performance is evaluated by the amount of rice deposited in the target bowl. Because the task is cyclic, the initial demonstrations are designed to start and end at approximately the same pose after one successful scoop.

\textbf{Pen-in-box}: The task is to pick up a randomly placed pen and place it into a cardboard box at a fixed location. A trial is counted as successful if the pen lands inside or remains on top of the box, without the pen being dropped or the box being moved significantly.

\textbf{Trousers Folding}: The task is to fold both trouser legs over the waist through two consecutive pick-and-place actions. Performance is scored per trial, with each fold contributing up to $0.5$: $0.25$ for a successful pick, requiring a stable grasp until the placement phase, and $0.25$ for a successful place, for a maximum total score of $1.0$.

\subsection{Datasets and Takeover Demonstrations}
For each task, we first collect an initial dataset $\mathcal{D}_I^{60}$ of $60$ expert demonstrations, varying object positions and orientations to increase diversity. From this dataset, we extract two smaller subsets: the first $20$ demonstrations, denoted $\mathcal{D}_I^{20}$, and the first $40$, denoted $\mathcal{D}_I^{40}$. We then train the corresponding policies $\pi_I^{20}$, $\pi_I^{40}$, and $\pi_I^{60}$. We then deploy $\pi_I^{20}$ and collect $20$ takeover demonstrations, denoted $\mathcal{D}_T^{20a}$, while varying object configurations. These are combined with the initial dataset $\mathcal{D}_I^{20}$ to form $\mathcal{D}_{IT}^{a}$, which is used to train a new policy, $\pi_{IT}^{a}$. Next, we deploy $\pi_{IT}^{a}$ and collect a further $20$ takeover demonstrations, denoted $\mathcal{D}_T^{20b}$. This yields the final dataset $\mathcal{D}_{IT}^{b} = \mathcal{D}_I^{20} \cup \mathcal{D}_T^{20a} \cup \mathcal{D}_T^{20b}$,
from which we train the final policy $\pi_{IT}^{b}$.

\begin{figure*}[t] \centering \includegraphics[width=0.99\linewidth]{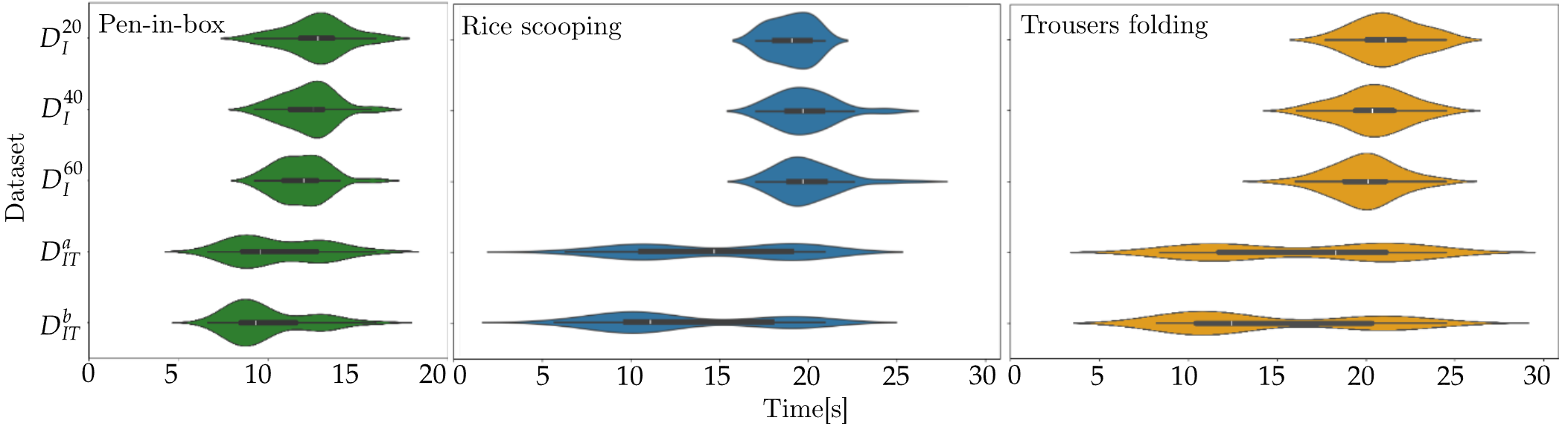} \caption{Lengths [s] of the demonstrations collected to train the different visuomotor policies for each task. The takeover paradigm produces significantly shorter demonstrations on average, highlighting its efficiency.} \label{fig:demo_length} \end{figure*}

\textbf{Analysis of Demonstrations Lengths.} We analyze the lengths of the demonstrations of each dataset used to train the different policies. Fig.~\ref{fig:demo_length} shows that the initial datasets $\mathcal{D}_I^{20}$, $\mathcal{D}_I^{40}$, and $\mathcal{D}_I^{60}$ have similar mean demonstration lengths across all tasks. In contrast, the takeover-based datasets $\mathcal{D}_{IT}^{a}$ and $\mathcal{D}_{IT}^{b}$ achieve lower mean demonstration lengths despite containing the same number of demonstrations. This reduction is most pronounced for \textit{Rice Scooping} and \textit{Trousers Folding}, but it is also present for \textit{Pen-in-box}. These results highlight the efficiency of the takeover paradigm in producing shorter, more targeted training data. By concentrating on failure cases encountered during deployment, takeover demonstrations improve policy robustness and reduce overall training time through their shorter duration.

\subsection{Experimental Evaluation}

We evaluate five policies ($\pi_I^{20}, \pi_I^{40}, \pi_{IT}^{a}, \pi_I^{60}, \pi_{IT}^{b}$) on each task. Each policy is tested over $10$ trials with different object positions and orientations, using the same $10$ task configurations across all policies to ensure a fair comparison. Videos of all experiments are available on the project website\footnote{\label{fn:website} \url{https://operator-takeover.github.io/}}.

\begin{figure}[t] \centering \includegraphics[width=0.99\linewidth]{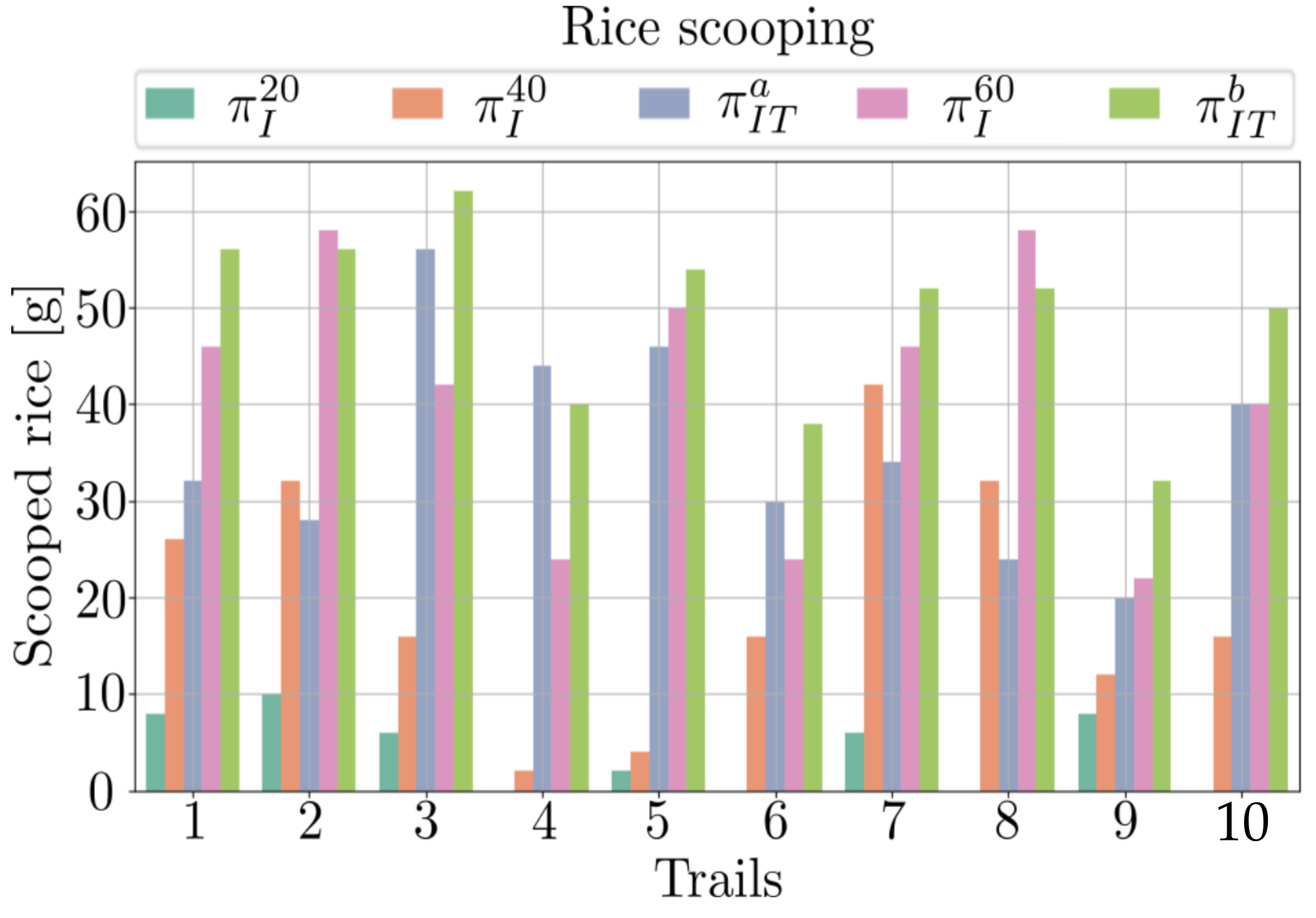} \caption{Detailed results of the cyclic rice scooping experiments. The amount of rice (in grams) is shown for each of the $10$ trials across all five evaluated policies.} \label{fig:results} \end{figure}

Results for \textit{Rice Scooping}, \textit{Pen-in-box} and \textit{Trousers Folding} are reported in Table~\ref{tab:scoop_mean},~\ref{tab:pen} and~\ref{tab:trousers}, respectively. For \textit{Pen-in-box} and \textit{Trousers Folding}, results are reported as the aggregate score across all trials in percentage, whereas for \textit{Rice Scooping} they are reported as the mean and standard deviation of the transferred rice mass in grams. Detailed per-trial results for \textit{Rice Scooping} are shown in Fig.~\ref{fig:results}.

\begin{table}[h!]
\centering
\begin{tabular}{|l|c|c|}
\hline
\textbf{Model}   & \textbf{\# Demos} & \textbf{Mean $\pm$ std [g]} \\ \hline \hline
$\pi_I^{20}$        & $20$                         & $4.0 \pm 4.0$                           \\ \hline \hline
$\pi_I^{40}$        & $40$                         & $19.8 \pm 12.9$                         \\ \hline
$\pi_{IT}^{a}$          & $40$                         & $\mathbf{35.4 \pm 11.0}$                         \\ \hline \hline
$\pi_I^{60}$         & $60$                         & $41.0 \pm 13.5$                         \\ \hline
$\pi_{IT}^{b}$          & $60$                         & $\mathbf{49.2 \pm 9.4}$                           \\ \hline
\end{tabular}
\caption{Means and Standard Deviations of the rice scooped in $45$ second over the $10$ trails.}
\label{tab:scoop_mean}
\end{table}

\begin{table}[h!]
\centering
\begin{tabular}{|l|c|c|c|}
\hline
\textbf{Model}   & \textbf{\# Demos} & \textbf{EE camera} & \textbf{Scene camera} \\ \hline \hline
$\pi_I^{20}$        & $20$                         & $30\%$       & $40\%$                    \\ \hline \hline
$\pi_I^{40}$        & $40$                         & $60\%$      & $40\%$                    \\ \hline
$\pi_{IT}^{a}$          & $40$                         & $\mathbf{100\%}$    & $\mathbf{90\%}$                     \\ \hline \hline
$\pi_I^{60}$         & $60$                         & $60\%$            & $70\%$              \\ \hline
$\pi_{IT}^{b}$          & $60$                         & $\mathbf{90\%}$       & $\mathbf{100\%}$                    \\ \hline
\end{tabular}
\caption{Aggregate score of Pen-in-box task over the $10$ trails.}
\label{tab:pen}
\end{table}

\begin{table}[h!]
\centering
\begin{tabular}{|l|c|c|c|}
\hline
\textbf{Model}   & \textbf{\# Demos} & \textbf{EE camera} & \textbf{Scene camera} \\ \hline \hline
$\pi_I^{20}$        & $20$                         & $30\%$       & $30\%$                    \\ \hline \hline
$\pi_I^{40}$        & $40$                         & $57.5\%$      & $45\%$                    \\ \hline
$\pi_{IT}^{a}$          & $40$                         & $\mathbf{70\%}$    & $\mathbf{70\%}$                     \\ \hline \hline
$\pi_I^{60}$         & $60$                         & $70\%$            & $70\%$              \\ \hline
$\pi_{IT}^{b}$          & $60$                         & $\mathbf{82.5\%}$       & $\mathbf{80\%}$                    \\ \hline
\end{tabular}
\caption{Aggregate score of Trousers folding task over the $10$ trails.}
\label{tab:trousers}
\end{table}

The baseline results show a clear improvement as the number of initial demonstrations increases. In \textit{Rice Scooping}, for instance, $\pi_I^{20}$ achieved an average of $4.00$ grams over $10$ trials, $\pi_I^{40}$ increased this to $19.80$ grams, and $\pi_I^{60}$ reached $41.00$ grams. A similar trend is observed in \textit{Pen-in-box} and \textit{Trousers Folding}, where performance generally rises from about $30\%$ to $70\%$ success as the number of demonstrations increases, for both end-effector and scene camera evaluations. These results underline the importance of larger demonstration datasets for improving baseline visuomotor policy performance.

To assess the effect of real-time operator takeover, we compare the baseline policy $\pi_I^{40}$ with the takeover-enhanced policy $\pi_{IT}^{a}$ for each task. In \textit{Rice Scooping}, $\pi_{IT}^{a}$ achieved $35.40$ grams on average, a $79\%$ improvement over $\pi_I^{40}$, while the final policy $\pi_{IT}^{b}$ outperformed $\pi_I^{60}$ by $20\%$, reaching $49.20$ grams per trial. In \textit{Pen-in-box}, the gains are even larger: $\pi_{IT}^{a}$ achieved $100\%$ and $90\%$ success using the end-effector and scene cameras, respectively, corresponding to an average improvement of $95\%$ over $\pi_I^{40}$. The final policy $\pi_{IT}^{b}$ also exceeded $\pi_I^{60}$ by about $46\%$ on average. The same trend appears in \textit{Trousers Folding}, where takeover-based policies again outperform the baselines, although with a smaller margin, probably due to the greater difficulty of the long-horizon deformable manipulation task. Finally, the difference between end-effector and scene camera inputs is generally small and does not exhibit a consistent pattern across these experiments, for either the baseline or takeover policies. This suggests that camera placement has a limited effect on overall performance, while takeover consistently improves results in all settings.

These results are particularly notable because the takeover datasets require substantially less demonstration time. In \textit{Rice Scooping}, for example, $\mathcal{D}_{IT}^{b}$ totals $465.6$ seconds, which is $46\%$ shorter than $\mathcal{D}_I^{60}$ at $869.0$ seconds. Despite this reduction, $\pi_{IT}^{b}$ outperforms the baseline trained on the full initial dataset, indicating that targeted takeover demonstrations provide more informative training data than additional full demonstrations.

\section{To Takeover or Not to Takeover}

In the current framework, the decision to initiate a takeover is made by the operator. We argue that this is a sensible approach, since the quality of the demonstrations strongly influences the quality of the learned policy~\cite{pari2021surprising}. At the same time, a more systematic way to estimate when the robot is approaching an undesirable state could further support intervention decisions.

Recall that our approach assumes that expert demonstrations correspond to desirable states. The DDPM policy is trained to match the conditional action distribution $p(A_t \mid O_t)$ induced by these demonstrations, where $A_t$ denotes actions and $O_t$ observations. As a result, policy performance can degrade when observations at inference time deviate substantially from the training distribution. Improving robustness to such shifts remains an active research problem in visuomotor diffusion policies~\cite{wang2023imitation, zhuang2024enhancing}.
This motivates the use of a metric that quantifies how far inference-time observations deviate from the training distribution, and that can serve as an out-of-distribution (OOD) signal to better inform takeover decisions.

\subsection{Mahalanobis Distance as OOD Metric}

The Mahalanobis distance measures how far a point lies from a distribution while accounting for correlations between variables:
\begin{equation}
d_M = \sqrt{(x - \mu)^\top \Sigma^{-1} (x - \mu)} .
\end{equation}
Here, $\mu$ denotes the mean, $\Sigma$ the covariance matrix, and $x$ the query point.
Because RGB images are high-dimensional, we compute embeddings of the images concatenated with the robot pose. Each observation is encoded into a $1056$-dimensional vector (two $512$-dimensional embeddings for the images and two $16$-dimensional embeddings for the pose). These embeddings serve as the conditioning input for the diffusion model.

\begin{figure}[t] \centering \includegraphics[width=0.99\linewidth]{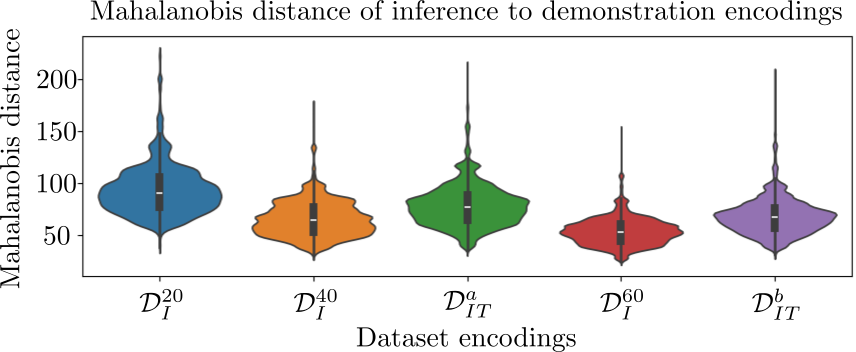} \caption{Violin plots of Mahalanobis distances for experimental observations compared to the learned distribution for each dataset (\textit{Rice scooping}).} \label{fig:m_distance_violine} \end{figure}

We evaluate the Mahalanobis distance in the \textit{Rice scooping} task, where we first encode all datasets $\mathcal{D}I^{20}$, $\mathcal{D}I^{40}$, $\mathcal{D}I^{60}$, $\mathcal{D}{IT}^a$, and $\mathcal{D}{IT}^b$ using their respective trained models ($\pi_I^{20}$, $\pi_I^{40}$, $\pi_I^{60}$, $\pi{IT}^a$, and $\pi_{IT}^b$). This produces embeddings $Z_I^{20}$, $Z_I^{40}$, $Z_I^{60}$, $Z_{IT}^a$, and $Z_{IT}^b$. Similarly, all observations recorded during the experiments ($\mathcal{E}$) are encoded using these models, resulting in $H_I^{20}$, $H_I^{40}$, $H_I^{60}$, $H_{IT}^a$, and $H_{IT}^b$.

\begin{figure*}[ht]
    \centering
    \includegraphics[width=0.99\linewidth]{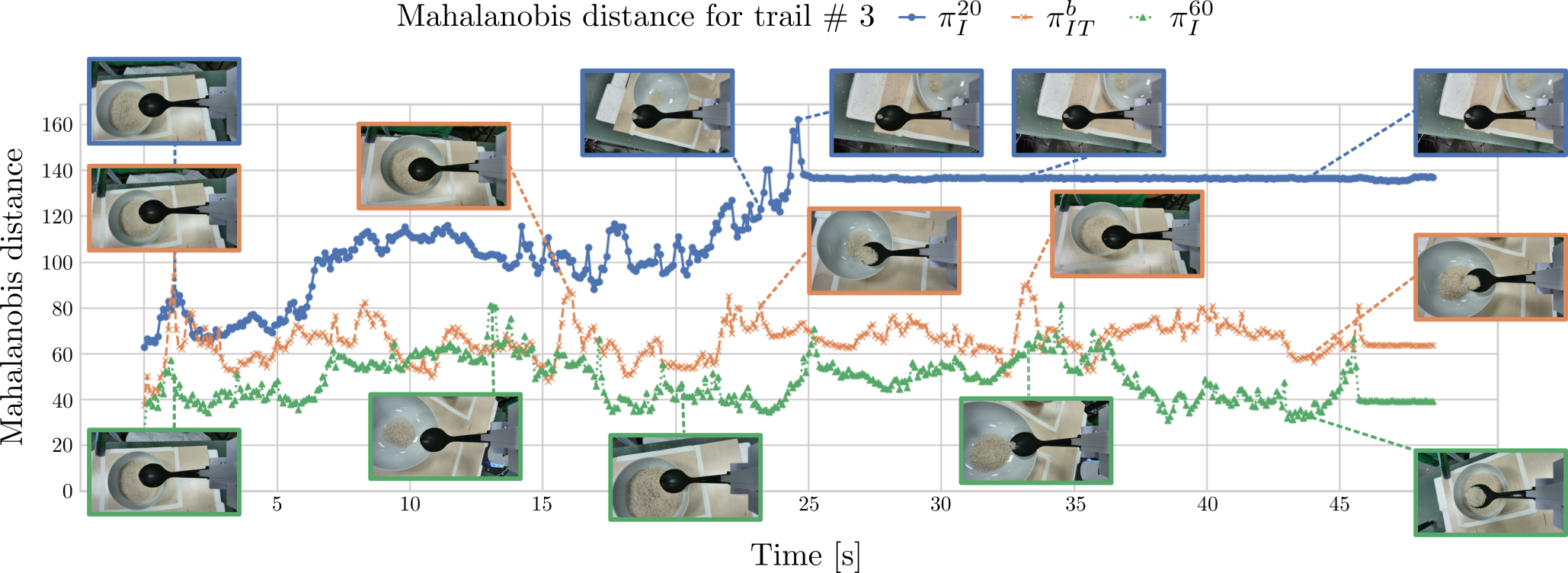}
    \caption{The Mahalanobis distance of the embedded observations of trail $\#3$ (\textit{Rice scooping}) of the models $\pi_I^{20}, \pi_{IT}^b,$ and $\pi_I^{60}$, with the RGB views shown for specific timesteps. }
    \label{fig:mdistancet03}
\end{figure*}

We estimate the mean ($\mu$) and covariance matrix ($\Sigma$) of each embedding set using the Minimum Covariance Determinant Estimator implemented in scikit-learn~\cite{scikit-learn}. For each encoded observation $h \in H$, the Mahalanobis distance $d_M$ is computed with Algorithm~\ref{alg:mdist}.

\begin{algorithm}
\caption{Mahalanobis Distance Calculation}
\label{alg:mdist}
\begin{algorithmic}[1]
    \State \textbf{Input:} Datasets $\mathcal{D}_I^{20}$, $\mathcal{D}_I^{40}$, $\mathcal{D}_I^{60}$, $\mathcal{D}_{IT}^a$, $\mathcal{D}_{IT}^b$, 
    \Statex \hspace{2em} trained models $\pi_{I}^{20}$, $\pi_{I}^{40}$, $\pi_{I}^{60}$, $\pi_{IT}^a$, $\pi_{IT}^b$, 
    \Statex \hspace{2em} experimental observations $\mathcal{E}$.
    \State \textbf{Output:} Mahalanobis distances $d_m$ for each observation encoding $h \in H$ with respect to each embedding set.

    \State \textbf{Step 1: Encode Dataset Embeddings} 
    \State     \[
    Z_I^{20} = \pi_{I}^{20}(\mathcal{D}_I^{20}), \quad Z_I^{40} = \pi_{I}^{40}(\mathcal{D}_I^{40}), \quad Z_I^{60} = \pi_{I}^{60}(\mathcal{D}_I^{60}), 
    \]
    \[
    Z_{IT}^a = \pi_{IT}^a(\mathcal{D}_{IT}^a), \quad Z_{IT}^b = \pi_{IT}^b(\mathcal{D}_{IT}^b)
    \]

    \State \textbf{Step 2: Encode Experimental Observations}
    \State     \[
    H_I^{20} = \pi_{I}^{20}(\mathcal{E}), \quad H_I^{40} = \pi_{I}^{40}(\mathcal{E}), \quad H_I^{60} = \pi_{I}^{60}(\mathcal{E}), 
    \]
    \[
    H_{IT}^a = \pi_{IT}^a(\mathcal{E}), \quad H_{IT}^b = \pi_{IT}^b(\mathcal{E})
    \]

    \State \textbf{Step 3: Calculate Mahalanobis Distances}
    
    \For{each  $Z \in \{Z_I^{20}, Z_I^{40}, Z_I^{60}, Z_{IT}^a, Z_{IT}^b\}$ and $H \in \{H_I^{20}, H_I^{40}, H_I^{60}, H_{IT}^a, H_{IT}^b\}$}
        \State $\mu_Z, \Sigma_Z = \texttt{MinCovDet}(Z)$
            \For{each encoding $h \in H$}
                \State
                $d_m(h, \mu_Z, \Sigma_Z) = \sqrt{(h - \mu_Z)^T \Sigma_Z^{-1} (h - \mu_Z)}$
            \EndFor
    \EndFor
\end{algorithmic}
\end{algorithm}

The summarized results of Algorithm~\ref{alg:mdist} are presented as violin plots in Fig.~\ref{fig:m_distance_violine}.
We observe that $\mathcal{D}_I^{20}$, which yields the highest average Mahalanobis distance, also corresponds to the worst-performing policy, $\pi_I^{20}$. This suggests that the Mahalanobis distance captures how far rollout observations deviate from the training distribution, and that adding demonstrations generally reduces this deviation. However, the relationship is not strictly monotonic: when comparing $\mathcal{D}_I^{40}$ with $\mathcal{D}_{IT}^{a}$ or $\mathcal{D}_I^{60}$ with $\mathcal{D}_{IT}^{b}$, similar average Mahalanobis distances can correspond to different task performance. For example, $\pi_{IT}^{b}$ outperforms $\pi_I^{60}$ in \textit{Rice Scooping} despite having a similar Mahalanobis distance profile.

\textbf{Case Study: Trial 3.}
To examine this further, we analyze Trial 3 in Fig.~\ref{fig:mdistancet03}, comparing the Mahalanobis distance profiles of $\pi_I^{20}$, $\pi_{IT}^{b}$, and $\pi_I^{60}$. The weakest policy, $\pi_I^{20}$, starts with the highest distance and reaches a peak of about $160$ around the $25$\,s mark, when the spoon moves outside the target bowl and the robot fails to recover, remaining stuck for the rest of the trial. In contrast, $\pi_{IT}^{b}$ and $\pi_I^{60}$ start from similar values (around $40$) and maintain substantially lower distances throughout the rollout.
Although $\pi_{IT}^{b}$ achieves the best task performance, its Mahalanobis distance profile remains broadly similar to that of $\pi_I^{60}$, indicating that the distance does not fully explain performance differences between strong policies. We also observe peaks near the end of the trial for both $\pi_{IT}^{b}$ and $\pi_I^{60}$, likely caused by abrupt robot stopping and the resulting shaking camera motion. More generally, the distance stays nearly constant when the visual input remains unchanged.

Overall, this case study suggests that the Mahalanobis distance is more useful for identifying undesirable states, such as failure or getting stuck, than for directly predicting final task performance. This supports its use as a signal for detecting critical failure points and motivating future automated recovery strategies.

\section{Conclusion}

In this work, we introduced a Real-Time Operator Takeover (RTOT) paradigm for training visuomotor diffusion policies through targeted human intervention during deployment. Instead of relying solely on increasingly large datasets of initial demonstrations, RTOT allows an operator to intervene only when the policy approaches failure, record the corrective segment together with its preceding context, and use this targeted data to iteratively retrain the policy. This turns deployment failures into supervision and provides a practical mechanism for improving robustness exactly where the policy is weakest.
Our experiments across three real-world manipulation tasks involving granular, rigid, and deformable objects show that this strategy consistently improves performance. Takeover-enhanced policies outperform baselines trained on equivalent numbers of initial demonstrations. Importantly, these gains are achieved with substantially shorter demonstrations, showing that targeted corrective data can be more informative than additional full trajectories. These results support a central conclusion of this work: in visuomotor imitation learning, demonstration relevance and coverage of failure cases can matter more than raw demonstration volume.

We also analyzed the Mahalanobis distance as a signal for identifying undesirable or out-of-distribution states during execution. Our results suggest that, while this measure does not fully predict final task performance, it is useful for highlighting critical failure states, such as situations in which the policy becomes stuck or deviates markedly from the training distribution. This makes it a promising ingredient for future systems that combine human takeover with automatic failure detection.

Overall, RTOT provides an efficient and adaptive framework for improving visuomotor policy training in real-world settings. By enabling seamless human intervention and converting corrective behavior into compact, high-value training data, it offers a practical path toward more robust and scalable robot learning for automating both domestic and industrial applications, including deformable object manipulation and repetitive pick-and-place tasks.

\section*{ACKNOWLEDGMENTS}

We used LLMs (GPT5) to assist with paper formatting and proofreading.

M.M. was funded by the European project euROBIN (grant No. 101070596).


\balance
\bibliographystyle{ieeetr}
\bibliography{references}

\end{document}